\documentclass[letterpaper, 10 pt, conference]{ieeeconf}  

\IEEEoverridecommandlockouts                              

\usepackage{graphics} 
\usepackage{epsfig} 
\usepackage{times} 
\usepackage{amsmath} 
\usepackage{amssymb}  
\usepackage{subcaption}
\usepackage{multirow}
\usepackage{cite}
\usepackage{url}
\usepackage{xcolor}

\title{\LARGE \bf
Leveraging Planar Regularities for Point Line Visual-Inertial Odometry\\
}

\author{Xin Li$^{*1, 2}$, Yijia He$^{*2}$, Jinlong Lin$^{1}$, Xiao Liu$^{2}$
\thanks{* Equal contribution, Xin Li's contribution was made when he was an intern in Megvii Research.}
\thanks{$^{1}$Xin Li and Jinlong Lin are School of Software \& Microelectronics, Peking University, Beijing, China,       
        \{\tt\small lixin97@pku.edu.cn, linjl@ss.pku.edu.cn\} }%
\thanks{$^{2}$Yijia He and Xiao Liu are with the Megvii (Face++) Technology Inc., Beijing, China,
	\{\tt\small heyijia2016@gmail.com, liuxiao@megvii.com\} }%
}

\begin{document}

\maketitle
\thispagestyle{empty}
\pagestyle{empty}

\begin{abstract}
With monocular Visual-Inertial Odometry (VIO) system, 3D point cloud and camera motion can be estimated simultaneously. Because pure sparse 3D points provide a structureless representation of the environment, generating 3D mesh from sparse points can further model the environment topology and produce dense mapping. To improve the accuracy of 3D mesh generation and localization, we propose a tightly-coupled monocular VIO system, PLP-VIO, which exploits point features and line features as well as plane regularities. The co-planarity constraints are used to leverage additional structure information for the more accurate estimation of 3D points and spatial lines in state estimator. To detect plane and 3D mesh robustly, we combine both the line features with point features in the detection method. The effectiveness of the proposed method is verified on both synthetic data and public datasets and is compared with other state-of-the-art algorithms.

\end{abstract}

\section{Introduction}

Simultaneous motion estimating and dense mapping are important to many robotic applications like autonomous driving, augmented reality, and building inspection. Camera and inertial measurement units (IMUs) are low-cost and effective sensors, based on which the  Visual-Inertial Odometry (VIO) system can be implemented. Existing VIO methods \cite{mourikis2007multi,qin2018vins,leutenegger2015keyframe} can estimate pose accurately but only provide a sparse map of 3D points. In \cite{pizzoli2014remode,newcombe2011dtam} dense mapping can be realized based on monocular vision and pixel-wise reconstruction, which is time-consuming and highly relies on Graphics Processing Unit (GPU) to guarantee real-time performance.

For light-weight dense mapping, Lucas et al.\cite{teixeira2016real} used sparse point features on image to generate 2D Delaunay triangulation and build a dense map for aerial inspection. 3D Delaunay triangulation, which is generated from sparse 3D points estimated by ORB-SLAM, is used to build a dense map in real-time for robot navigation \cite{ling2017building}. However, these algorithms decouple pose estimation and dense mapping, causing information loss. Antoni et al. \cite{rosinol2019incremental,rosinol2019master} proposed an incremental visual-inertial 3D mesh generation system that uses 3D mesh to detect plane and enforces planar structural regularities to improve localization accuracy. However, the spares points-based map is structureless so that the mesh does not fit the environment well.

\begin{figure}[thpb]
	\centering
	\includegraphics[width=0.45\textwidth]{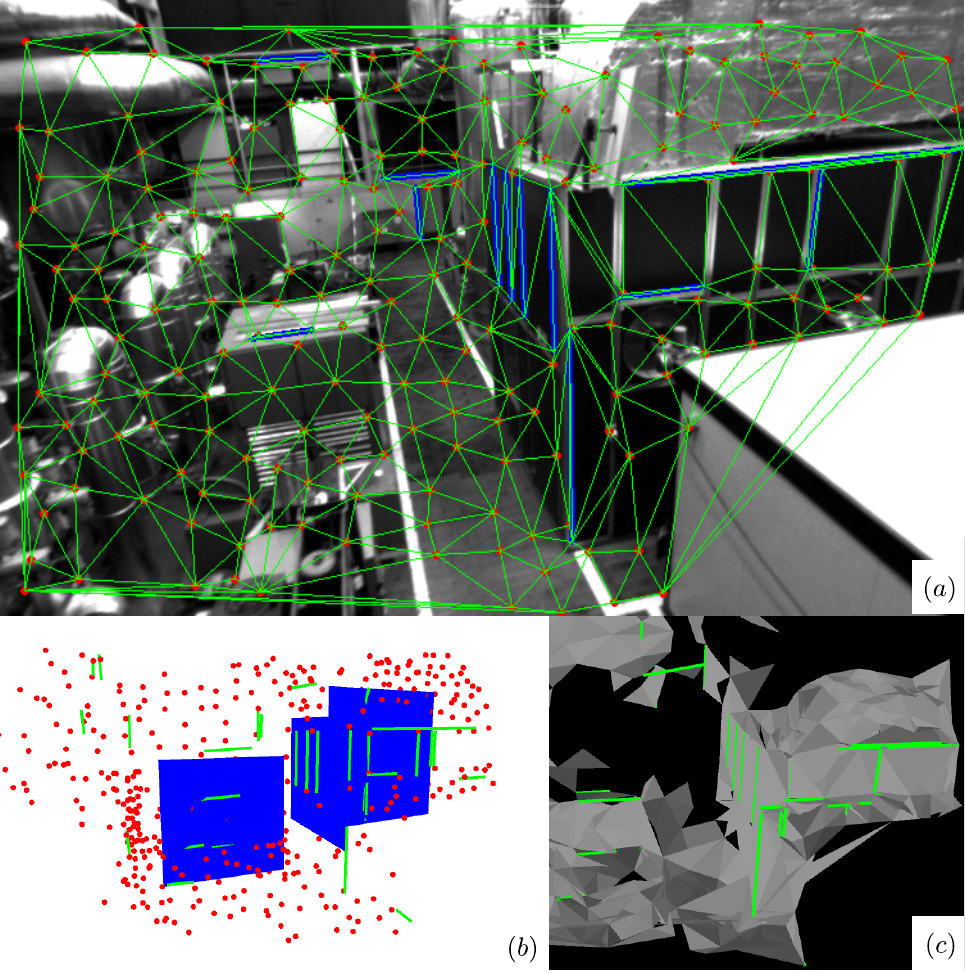}
	\caption{The proposed monocular VIO system that detects planes and builds 3D mesh of the environment by combining 3D sparse points and structural lines. (a) 2D Delaunay triangulation of point features and line features. (b) Plane detection (blue) with the 2D triangulation result and the 3D structural line (green). (c) 3D mesh of the environment and the related structural lines (green). }
	\label{fig:show_contri}
\end{figure}

For heterogeneous features utilization, the line and plane feature have attracted researchers' attention. In our previous work \cite{he2018pl}, The environment reconstructed with 3D straight lines can reflect more geometric structure information than point features. Moreover, line features incorporated into the SLAM system can improve pose accuracy and the robustness under illumination-changing scenes \cite{he2018pl,pumarola2017pl,zuo2017robust}. However, they do not consider the geometric constraint between point and line features. Structural regularity from the Atlanta world model is introduced into the line based visual odometry system \cite{zou2019structvio, li2019leveraging}. But point features are not considered in their system. Moreover, all these works did not use structural landmarks to build a dense map.
Planar constraints used in these SLAM system are more often relies on the depth sensor\cite{ataer2013tracking,kaess2015simultaneous,yang2019tightly,zuo2019lic}.
Lu et al.\cite{lu2015visual} presented a multi-layer feature graph utilizing points, lines, planes, and vanishing points to improve the accuracy of bundle adjustment. However, they used RANSAC-based approaches to detect planes which require numerous iteration and are time-consuming. Yang et al.\cite{yang2018observability} investigated the observability of the VIO using heterogeneous features of points, lines, and planes.
\cite{nardi2019unified} proposed a unified representation for all the point, line and plane features but they need depth sensors. 

In this paper, we propose a tightly-coupled monocular VIO system that involves heterogeneous visual features including points, lines, and planes as well as their co-planarity constraints (PLP-VIO), to realize accurate motion estimation and dense mapping, as shown in Fig.\ref{fig:show_contri}. A related work \cite{rosinol2019incremental} also utilized planar regularities, which only used sparse point features. In comparison, our proposed method not only leveraged the planar regularities but also involves both point features and line features to detect planes, so that richer structural information is used for 3D mesh generation. The main contributions of this work include:

\begin{itemize}
	\item We propose a non-iterative plane detection method and a 3D mesh generation method based on sparse points and spatial lines for monocular VIO. Our proposed method utilizes structural line to improves the correctness and robustness of plane detection and mesh generation compared to the method involving only sparse points\cite{rosinol2019incremental}.
		
	\item We develop a tightly-coupled monocular VIO system that utilizes heterogeneous visual features, include points, lines, and planes, as well as their co-planarity constraints. The richer visual information and spatial constraints between landmarks improve the estimator accuracy and robust.
	   
	\item We provide an ablation experiment to verify the effectiveness of heterogeneous features on both synthetic data and the \texttt{EuRoc} dataset \cite{burri2016euroc}. Experimental results demonstrated that our system simultaneously improves the accuracies of pose estimation and mapping.
\end{itemize}

\section{System Overview}
\begin{figure}[thpb]
	\centering
	\includegraphics[width=0.47\textwidth]{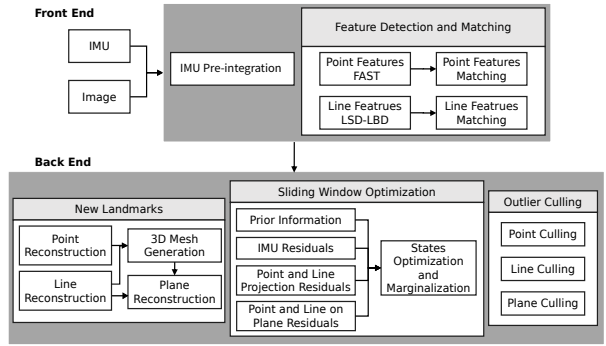}
	\caption{Overview of our PLP-VIO system. The Front-End module is used to extract information from the raw measurements; The 3D points, lines, mesh map, and the body states are estimated with sliding window optimization in the Back-End.}
	\label{fig:sys_overview}
\end{figure}

The proposed system is derived from our previous work PL-VIO\cite{he2018pl} which incorporate line segments into Vins-Mono\cite{qin2018vins}.
As shown in Fig. \ref{fig:sys_overview}, our system contains two modules: the front end and the back end. In the front end, raw measurements from IMU and camera are pre-processed. The related operations include IMU pre-integration, detection, and matching of point-line features, which are described in \cite{he2018pl}. Here we mainly introduce the plane-related operations in the back end. 

First, the 2D point and line features are triangulated to estimate the 3D point and line landmarks in the map. Second, these 3D landmarks are used to generate plane and 3D mesh. Also, co-planar constraints are extracted from the planes to constraint the 3D landmarks. A detailed description of the plane reconstruction will be presented in Sec. \ref{sec:3Dmeshcoplanar}. Third, the IMU body state and 3D landmarks in the map will be optimized with the sliding window optimization by minimizing the sum of the IMU residuals, the vision re-projection residuals, and the co-planar constraints residuals. All these residuals will be described in Sec. \ref{sec:method}. The marginalization strategy of the keyframe in the sliding window can be found at \cite{qin2018vins}. Finally, the invalid planes are culled. If active point and line features belonging to a plane are fewer than the threshold number $\delta_\pi = 30$, this plane is culled from the map and the co-planar constraints with this plane are also removed.

\section{Co-planarity Detection} 
\label{sec:3Dmeshcoplanar}

Given a set of 3D points and spatial lines, the planes and their co-planarity are detected. Below we introduce the landmarks representation, the 3D mesh generation from point-line features, and plane detection with sparse landmarks sequentially.

\subsection{Landmarks Representation}
Since 3D plane detection is based on the 3D landmarks representation, here we briefly introduce the geometry meaning and parameterization of these 3D landmarks.

\subsubsection{Point representation}
We use the inverse depth $\lambda \in \mathbb{R}$ to parameterize the point landmark from the first keyframe in which this point is observed. Given its observation $\mathbf{z} =[u,v,1]^{\top}$ in the normalized image plane $\mathbf{I}$ (located at the focal length $=1$), we can get its 3D position by $\mathbf{f} = \frac{1}{\lambda} \cdot \mathbf{z}$. 

\subsubsection{Line representation}
Pl\"ucker coordinates $\mathcal{L}^{c}=\left[\mathbf{n}^{c\top},\mathbf{d}^{c\top}\right]^{\top} $ are used for 3D line initialization and transformation.  $\mathbf{d}^{c} \in \mathbb{R}^3$ is the line's direction vector in camera frame $c$, and $\mathbf{n}^{c} \in \mathbb{R}^3$ is the normal vector of the plane determined by this line and the camera frame's origin point. The Pl{\"u}cker coordinates are over-parameterized since there is an implicit constraint between the vector $\mathbf{n}$ and $\mathbf{d}$, i.e., $\mathbf{n}^{\top}\mathbf{d}=0$. The orthonormal representation $\mathcal{O}=(\mathbf{U},\mathbf{W})\in SO(3)\times SO(2)$ is used in optimization to avoid over-parameterization. The geometric representation of 3D line and the transformation between parameterized forms are introduced in detail at \cite{zhang2015building}. 

\subsubsection{Plane representation}
\begin{figure}[thpb]
	\centering
	\includegraphics[width=0.25\textwidth]{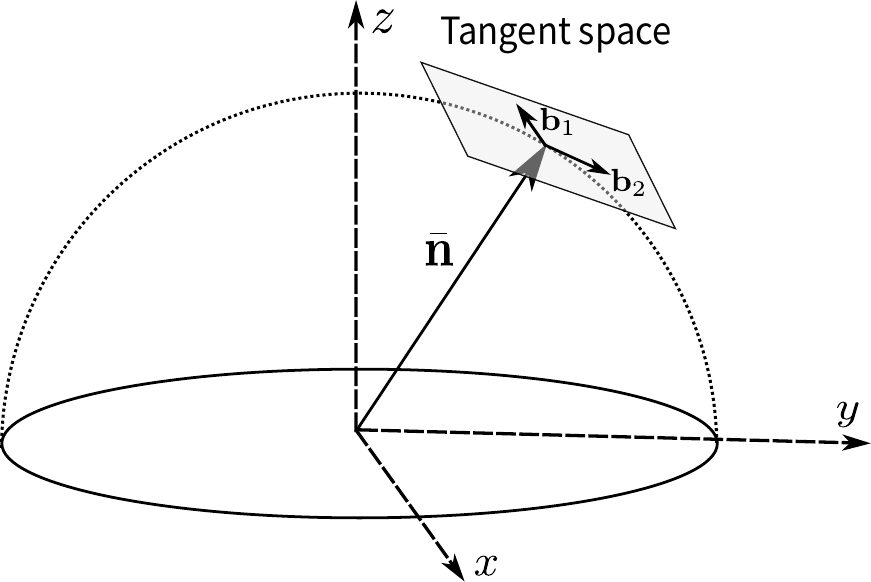}
	\caption{Illustration of the parameterization of the 2 DoF  normal vector with two orthogonal basis $\mathbf{b}_1, \mathbf{b}_2$, which span the tangent space.}
	\label{fig:tangent space}
\end{figure}
We denote a plane in world frame $w$ by $\boldsymbol{\pi} = \left[\mathbf{n}^{\top},{d}^{\top}\right]^{\top}$, where $\mathbf{n} \in \mathbb{R}^3$ is the plane's normal vector, $ d \in \mathbb{R}$ is the distance from frame origin to plane. The unit normal vector $\mathbf{n}$ have three parameters but only two Degrees of Freedom (DoF). We therefore optimize the plane normal vector $\mathbf{n}$ on its tangent space during optimization. Inspired by the gravity vector's parameterization in \cite{qin2018vins}, we represent $\mathbf{n}$ as $\bar{\mathbf{n}} + w_1 \mathbf{b}_1 + w_2 \mathbf{b}_2$, where $ \bar{\mathbf{n}} $ is a unit normal vector. $ \mathbf{b}_1 $ and $ \mathbf{b}_2 $  are two orthogonal basis spanning the tangent plane, as shown in Fig.\ref{fig:tangent space}.

\subsection{3D Mesh Generation}

\begin{figure}[thpb]
	\centering
	\includegraphics[width=0.32\textwidth]{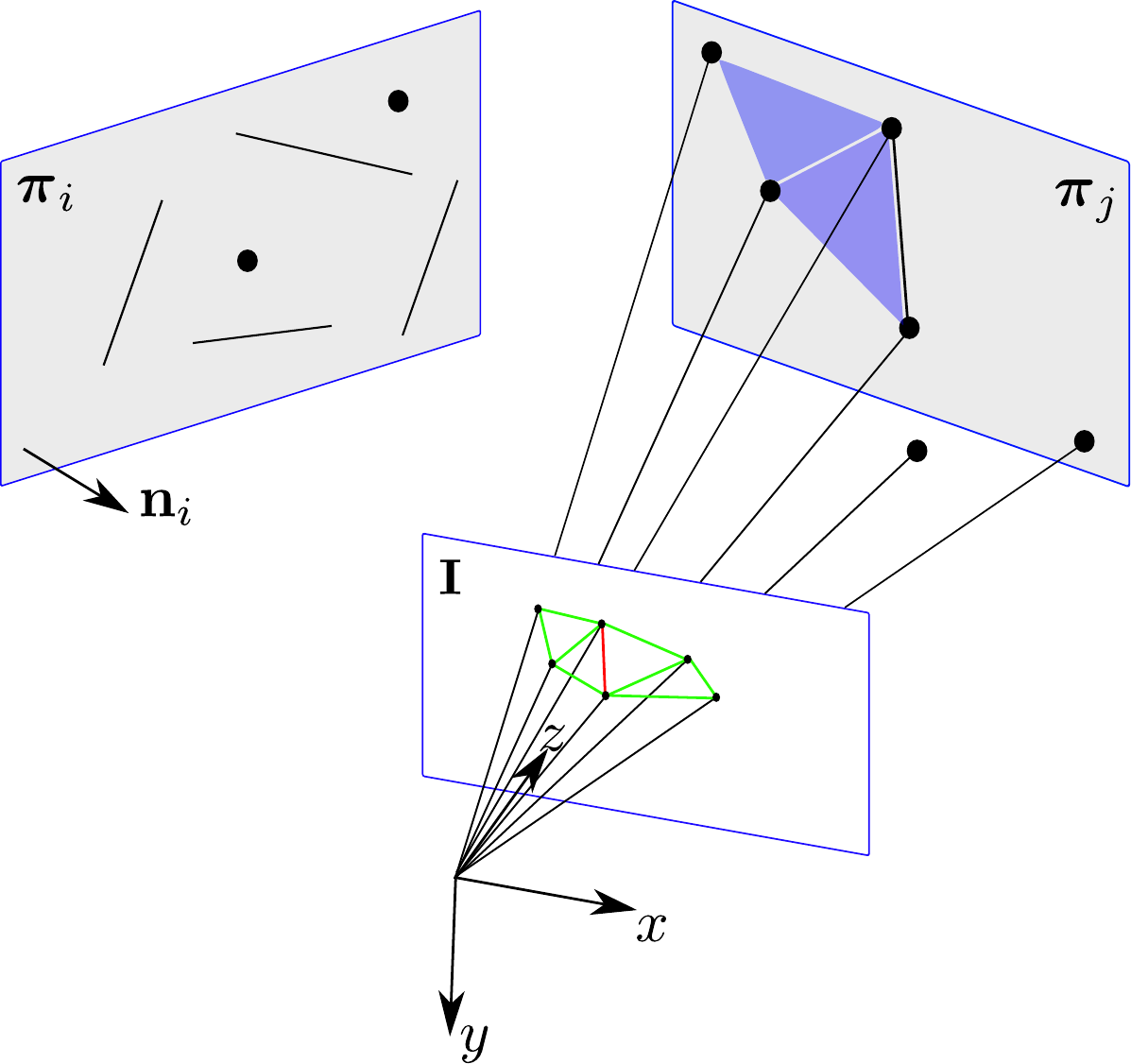}
	\caption{3D mesh generation and plane detection. 2D Delaunay triangulation (green) build with point (black dot) and line (red) features in image plane $\mathbf{I}$ are used to build 3D mesh (blue patches).  Plane $\boldsymbol{\pi}_j$ is detected with 3D mesh and spatial lines. Plane $\boldsymbol{\pi}_i$ is detected mainly using spatial lines when point features are few.}
	\label{fig:mesh_plane}
\end{figure}

Since it is difficult to directly generate 3D mesh from sparse 3D landmarks, we build 3D mesh from 2D Delaunay triangulation using both point and line features, as shown in Fig.\ref{fig:mesh_plane}. Firstly, we detect point and line features in the latest keyframe. Then, we use the constrained Delaunay triangulation algorithm \cite{chew1989constrained} to build a 2D mesh from these points and line segments. Compared to the basic Delaunay triangulation, the constrained Delaunay triangulation retains the observed line edges on the image frame. Thus, the structural information of line features is preserved. 
 
However, there might be some invalid 3D triangular patches since they build from 2D observation. To remove these outliers, we filter the patches according to two conditions: if a 3D triangular patch is on a plane, there must be more than two adjacent patches that the angle between their normal vectors is less than a certain threshold, such as $5$deg. The other one condition is used in \cite{rosinol2019incremental} that 3D triangular patches should not be particularly sharp triangles, such as triangles with the aspect ratio higher than $20$ or an acute angle smaller than $5$deg.

We maintain the update of the 3D mesh map over the sliding window. When the 3D points and lines are optimized during the optimization, each triangular patch will also be updated. When 3D points and lines are marginalized, the associated mesh will be fixed. Considering the instability of the straight-line endpoint detection, we do not directly propagate the mesh between frames instead of detecting mesh on every new frame and merging it into the local map.
Mesh fusion is conducted by comparing the current 3D triangular patches to the previous 3D mesh in the sliding window and removing duplicated patches. If the state estimator does not provide any 3D points or lines, which might happen during fast rotation, the 3D mesh will not be generated. 

\subsection{Plane Detection}
\label{sec:pd} 

After getting the 3D mesh in the scene, we can detect planes from these triangular patches and spatial lines as shown in Fig.\ref{fig:mesh_plane}. Utilizing both triangular patches and all spatial lines to detect planes is more suitable for not only point-rich but also textureless scenes, compared to utilizing only triangular patches or only lines. 


Before finding the plane, we collect useful information from triangular patches and spatial lines. For each triangular patch, we compute its normal direction $\mathbf{n}_\pi$. For spatial lines, the pl\"ucker coordinates treat a 3D line as an infinite straight line, and we select two 3D points on the line to help to detect plane. Since the observation of a straight line in an image is a line segment, the endpoints of a line segment can be used to calculate the two 3D points on a spatial line \cite{zhang2015building}. Moreover, the reconstructed 3D lines have noise, we need to filter or merge some 3D lines. If two lines have similar directions $\mathbf{d}$ and are close enough in 3D space, we will regard them as the same straight line in the process of plane detection. 

In order to extract planes in a non-iterative manner, we divide the planes in space into horizontal planes, vertical planes, and other planes. The horizontal and vertical planes can be identified using the direction of gravity measured by IMU measurements.
 
As to horizontal plane detection, we collect all triangular patches and spatial lines that are vertical to the gravity direction. Then, we build a 1D histogram of the height of vertices of triangular patches and the height of 3D points on lines. 
Moreover, considering the structural information of lines, we give twice the weight to 3D points on the lines than to other points. After statistics, a Gaussian filter is used to eliminate multiple local maximums as used in \cite{rosinol2019incremental}. Finally, we extract the local maximum of the histogram which exceeds the threshold ($\sigma_t = 20$) and considers it a valid plane.

As to vertical plane detection, A 2D histogram is built, whose one axis is the azimuth $\theta$ of the plane normal vector, and the other axis is the distance from the origin to the plane. The histogram consists of $n_\theta \times n_d$ bins with an bin size $\delta \theta \times \delta d$. For triangular patches, only those patches that are perpendicular to a horizontal normal vector $\mathbf{n}\left(\theta_i\right)=\left[\cos\theta_i,\sin\theta_i,0\right]^{\top}$ is used for statistics. If a vertical patch with azimuth $\theta_i$ and distance $d_j$, then the number in histogram bin $( \lfloor\frac{\theta_i}{\delta \theta}\rfloor, \lfloor\frac{d_j}{\delta d}\rfloor)$ will be increased by 1, where $\lfloor\cdot\rfloor$ is floor operator. Line statistics are more complicated. We need to classify straight lines into two types: parallel or non-parallel to gravity. For lines which are non-parallel to gravity, a horizontal normal vector is obtained by cross producting the gravity direction vector and the line's direction vector, the azimuth of the horizontal normal vector is used for statistics. The projection distance from a 3D point on the line onto the horizontal normal vector is used as the distance statistic. For lines which are parallel to gravity, since it is perpendicular to all horizontal normal vectors, we need to traverse to calculate the projection distance of a 3D point on the line onto all horizontal normal vectors with azimuth $\theta_i = i * \delta \theta,i = 0,1...,n_\theta - 1$. 
After statistics, we select the candidate planes in a similar way of horizontal plane detection.

As for other planes, the above non-iterative method cannot be adopted since the angle of the normal vector have 2 DoF. Considering the efficiency of plane detection, these planes are not detected in our system. Fortunately, a lot of patches and lines have been classified after detecting the horizontal and vertical planes in a man-made structural environment.


After plane detection, we need to detect whether a plane is already in the estimator according to their parameters $\boldsymbol{\pi}$. If close enough they will be removed to avoid duplicated plane variables. To generate planar regularity to improve the VIO accuracy, we will assign the 3D points and spatial lines to their corresponding plane when the distance from landmark to plane is lower than a threshold. After the sliding window optimization, if the distance from the point or the line endpoints to the plane exceeds $ 3 $cm, then the point or line will not be constrained by the planar regularity, and meshes which use these points or lines as vertices will be removed.

\section{PL-VIO With Planar Constraint} 
\label{sec:method}
   
In this section, we build our VIO system with the sliding window optimization which estimates body states and 3D landmarks by fusing IMU and visual information. 

\subsection{Formulation}

The full state vector of the proposed VIO system within a sliding window contains the IMU states, point features' depths, 3D line landmarks, and plane states. The states at time $t$ are defined as:
\begin{equation}
\begin{aligned}
\boldsymbol{\mathcal{X}}_t & \doteq  \{\mathbf{x}_i, \lambda_k, \mathcal{O}^{w}_l, \boldsymbol{\pi}^{w}_h \}_{i\in\mathcal{B}_t, k \in \mathcal{F}_t, l \in \mathcal{L}_t, h \in \Pi _t} \\
\mathbf{x}_{i} & \doteq \left[ \mathbf{p}_{wb_i}, \mathbf{q}_{wb_i}, \mathbf{v}^{w}_{b_i}, \mathbf{b}^{b_i}_{a}, \mathbf{b}^{b_i}_{g} \right]
\end{aligned}
\end{equation}
where the set $\mathcal{B}_t$ contains the active IMU body states in the sliding window at time $t$. The sets $\mathcal{F}_t$, $\mathcal{L}_t$ and $\Pi_t$ contain the point, line and plane states active in sliding window at time $t$, respectively. $\mathbf{x}_i$ is composed of $i^{th}$ IMU body position $\mathbf{p}_{wb_i} \in \mathbb{R}^3 $, orientation $ \mathbf{q}_{wb_i}$, velocity $\mathbf{v}^{w}_{b_i}\in \mathbb{R}^3$ with respect to the world frame $w$. $\mathbf{b}^{b_i}_{a},\mathbf{b}^{b_i}_{g} \in \mathbb{R}^3$ are accelerometer bias and gyroscope bias in body frame $b_i$, respectively. We use quaternion to represent orientation within state vector and use rotation matrix $\mathbf{R} \in  SO(3)$ to rotate vectors. For landmarks, We use the inverse depth $\lambda_k $ to parameterize the $k^{th}$ point landmark from the first keyframe in which the landmark is observed. $\mathcal{O}^{w}_l$ and $\boldsymbol{\pi}^{w}_{h}$ are the orthonormal representations of the $l^{th}$ line feature and the $h^{th}$ plane in the world frame $w$, respectively. 

\begin{figure}[t]
	\centering
	\includegraphics[width=0.4\textwidth]{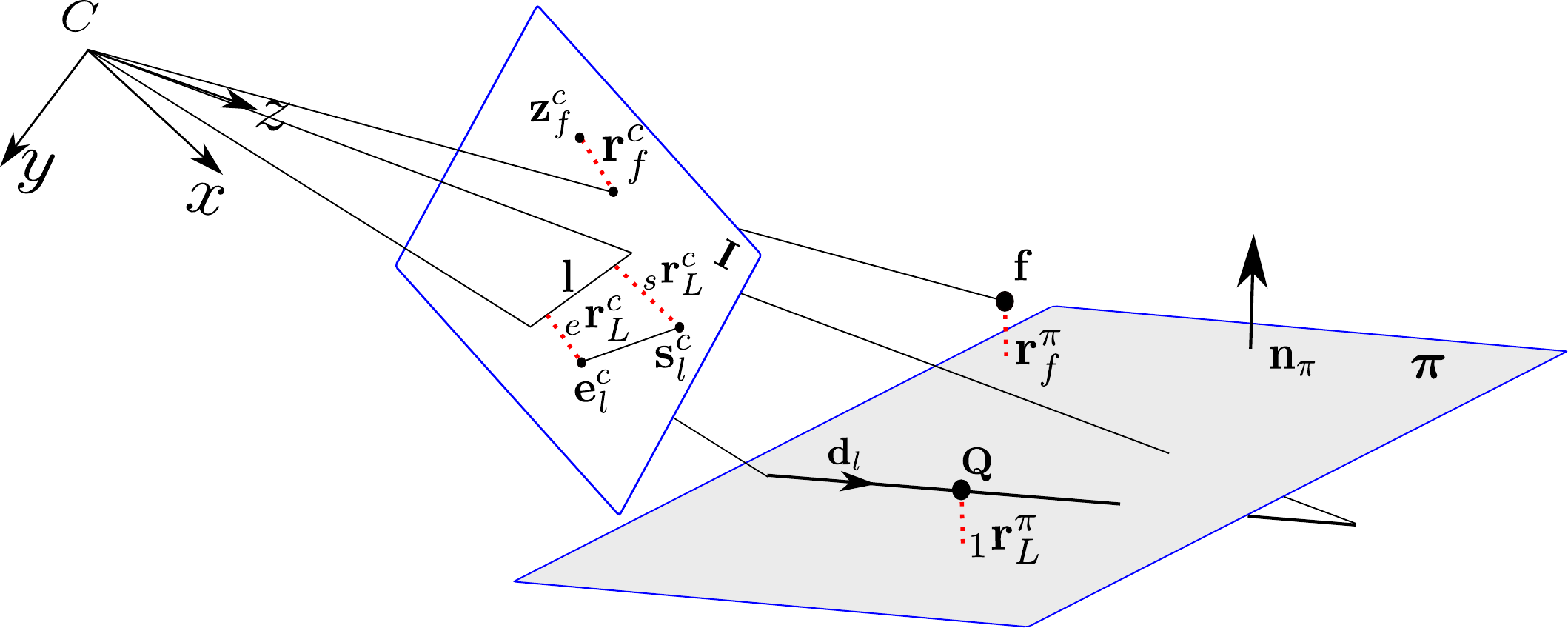}
	\caption{ Illustration of visual residuals (red dash line) : point re-projection residual $\mathbf{r}^c_f$, line re-projection residual $\mathbf{r}^c_L$, the distance from 3D point to plane $\mathbf{r}^\pi_f$, the distance from 3D line to plane $\mathbf{r}^\pi_L$.}
	\label{fig:resdiuals}
\end{figure}

We optimize all the state variables in the sliding window by minimizing the sum of cost terms from all the measurement residuals:
$$
\begin{aligned}
\boldsymbol{\mathcal{X}} = & \arg \min _{\boldsymbol{\mathcal{X}}} \left\|\mathbf{r}_{p} - \mathbf{H}_p\mathcal{X} \right\|^{2} +\sum_{i \in \mathcal{B}} \left\|\mathbf{r}_{b}\right\|_{\Sigma_{b_{i} b_{i+1}}}^{2} \\
&+\sum_{i \in \mathcal{B}}\left(\sum_{k \in \mathcal{F}} \rho\left(\left\|\mathbf{r}^{c_i}_{f_{k}}\right\|_{\Sigma_{\mathcal{F}}}^{2}\right) + \sum_{l \in \mathcal{L}} \rho\left(\left\|{\mathbf{r}^{c_i}_{L_l}}\right\|_{\Sigma_{\mathcal{L}}}^{2}\right) \right) \\
&+\sum_{h \in \Pi} \left( \sum_{ k \in \mathcal{F}} \rho\left(\left\|\mathbf{r}^{\pi_h}_{f_k}\right\|_{\Sigma_{\mathcal{F}}^{\pi}}^{2}\right)+\sum_{l \in \mathcal{L}} \rho\left(\left\|{\mathbf{r}^{\pi_h}_{L_l}}\right\|_{\Sigma^{\pi}_{\mathcal{L}}}^{2}\right) \right)
\end{aligned}
$$
where $\mathbf{r}_b$ is the IMU measure residuals. $\mathbf{r}^{c_i}_{f_{k}}$ and $\mathbf{r}^{c_i}_{L_l}$ are
re-projection residual of point and line, respectively. $\mathbf{r}^{\pi_h}_{f_k}$ and $\mathbf{r}^{\pi_h}_{L_l}$ are point-on-plane residual and line-on-plane residual, respectively. An illustration of these residuals is shown in Fig.\ref{fig:resdiuals}. $\mathbf{r}_p$ and $\mathbf{H}_p$ are the prior residual and Jacobian from marginalization operator, respectively \cite{qin2018vins}. Since the IMU pre-intergration residual and sliding window marginalization is not our main contribution, their details can be found in \cite{qin2018vins}.  $\rho\left(\cdot\right)$ is the Cauchy robust funtion used to suppress outliers. $\boldsymbol{\Sigma}_{(\cdot)}$ is the covariance matrix of a measurement. The covariance matrices of measurements are constant in our experiments except $\Sigma_{b_ib_{i+1}}$, which is calculated by covariance propagation with IMU measurement noise. The setting of each covariance will be introduced in the subsequent sections. The Ceres solver \cite{ceres-solver} is used to solve this nonlinear problem.

%
%

\subsection{Point Feature Measurement Model}  

For point features, the re-projection error of a 3D point is the distance between the projected point and the observed point in the normalized image plane. Given the $k^{th}$ point feature measurement at frame $c_j$, $\mathbf{z}_{f_k}^{c_j}=[u_{f_k}^{c_j},v_{f_k}^{c_j},1]^{\top}$, the re-projection error is defined as:

\begin{equation}
\label{eq:point_error}
\begin{split}
\mathbf{r}^{c_i}_{f_{k}} &= \begin{bmatrix}
\frac{x^{c_j}}{z^{c_j}} - u_{{f}_k}^{c_j} \\ \frac{y^{c_j}}{z^{c_j}} -v_{{f}_k}^{c_j}
\end{bmatrix} \\
\mathbf{f}_k^{c_j}=\begin{bmatrix}
x^{c_j}\\y^{c_j}\\z^{c_j}
\end{bmatrix}&= \mathbf{R}^{\top}_{bc}(\mathbf{R}^{\top}_{wb_j}(\mathbf{R}_{wb_i}((\mathbf{R}_{bc}\frac{1}{\lambda_k}\begin{bmatrix}
u_{{f}_k}^{c_i} \\ v_{{f}_k}^{c_i} \\ 1
\end{bmatrix} \\& + \mathbf{p}_{bc}) + \mathbf{p}_{wb_i}) - \mathbf{p}_{wb_j})-\mathbf{p}_{bc})
\end{split}
\end{equation}
where $\mathbf{z}_{{f}_k}^{c_i}=[u^{c_i}_{{f}_k},v^{c_i}_{{f}_k},1]^{\top}$ is the first observation of the feature in camera frame $c_i$. $\mathbf{R}_{bc}$ and $\mathbf{t}_{bc}$ is the extrinsic parameters between the camera frame and the IMU body frame. The setting of the covariance matrix $\boldsymbol{\Sigma}_{\mathcal{F}}$ is based on the assumption that the feature point coordinates are affected by isotropic white Gaussian noise. The standard deviation of the noise is set with 1 pixel.
 
\subsection{Line Feature Measurement Model}

Usually, the re-projection error of a line segment is defined as the distance from its endpoints to the projected line in the normalized image plane. However, the observation of the same spatial line in different frames have different lengths. Therefore, we normalize the distance with the line segment lengths as our line re-projection residuals.
 
Given a 3D line with its Pl\"ucker coordinates in camera frame $\mathcal{L}^{c}_l=\left[\mathbf{n}^c_l,\mathbf{d}^{c}_l\right]$. The measurement of the line segment $\mathbf{z}^{c}_{L_l}$ in the normalized image plane consists with two endpoints $\mathbf{s}^{c}_l=[u_s,v_s,1]^{\top}$ and $\mathbf{e}^{c}_l=[u_e,v_e,1]^{\top}$. The line re-projection residual is defined as:
\begin{equation}
\label{eq:errL}
\mathbf{r}^{c}_{L_l} =\frac{1}{\left\| \mathbf{z}^{c}_{L_l} \right\|}\begin{bmatrix}
d(\mathbf{s}_{l}^{c},\mathbf{n}_l^{c}) \\ d(\mathbf{e}_l^{c},\mathbf{n}_l^{c})
\end{bmatrix}
\end{equation}
with $d(\mathbf{s},\mathbf{n})$ indicating the distance function from endpoint $\mathbf{s}$ to the projection line $\mathbf{l}$:

\begin{equation}
\label{eq:dis}
d(\mathbf{s},\mathbf{n}) = \frac{\mathbf{s}^\top\mathbf{n}}{\sqrt{n_1^2+n_2^2}}
\end{equation}
For the setting of covariance $ \boldsymbol{\Sigma}_{\mathcal{L}} $, we assume that the end-point of the line segment is also affected by white Gaussian noise as with point features.
\subsection{Plane Feature Measurement Model}

3D landmarks co-planar constraints are defined by the distances from 3D landmarks to the 3D plane.

\subsubsection{Point on plane residual}Given a 3D point $\mathbf{f}_k$ on the plane $\boldsymbol{\pi}_h$, the residual is defined as:
\begin{equation}
\mathbf{r}^{\pi_h}_{f_k} = \mathbf{n}_\pi \cdot \mathbf{f}_k - d_\pi
\end{equation}

\subsubsection{Line on plane residual}

A line on the plane can provide information from two aspects\cite{nardi2019unified}. 
\begin{itemize}
	\item every point on this spatial line must be on the plane.
	\item the direction of this line and the plane normal vector are orthogonal.
\end{itemize}

Hence, given 3D lines Pl\"ucker coordinates  $\mathcal{L}_l = \left[ \mathbf{n}_l, \mathbf{d}_l\right]^\top$, 
the residual is given by:

\begin{equation}
\begin{aligned}
_1\mathbf{r}^{\pi_h}_{L_l} &= \mathbf{n}_\pi \cdot \mathbf{Q} - d_\pi \\
_2\mathbf{r}^{\pi_h}_{L_l} &= \mathbf{n}_\pi \cdot \mathbf{d}_l 
\end{aligned}
\end{equation}
where $\mathbf{Q} = \frac{\mathbf{n}_l \times \mathbf{d}_l}{||\mathbf{d}_l||}$ corresponds to a point in the 3D line.

According to the plane detection in Sec.\ref{sec:pd}, the residual error of the distance from the point to the plane will not exceed $3$cm, we model the distance residual with white Gaussian noise with the standard deviation is 1cm. As for the angle uncertainty of $ _2\mathbf{r}^{\pi_h}_{L_l} $, the standard deviation is set with 1deg.
\section{EXPERIMENTAL RESULTS} \label{expentimental}

To evaluate the proposed approach we conduct experiments on synthetic data and the \texttt{EuRoc} dataset. The trajectory accuracy, mapping quality, and runtime are analyzed. An intuitive demonstration can be viewed in the supplementary video\footnote{\url{https://www.youtube.com/watch?v=V1KU6V49UKI}}. In these experiments, the algorithm ran on a computer with Intel Core i7-8550U @ 1.8GHz, 16GB memory and the ROS Kinetic\cite{quigley2009ros}.

To assess the advantages of our proposed approach, we compared our method with OKVIS with monocular mode \cite{leutenegger2015keyframe}, VINS-Mono \cite{qin2018vins} without loop closure, PL-VIO \cite{he2018pl}, PP-VIO \cite{rosinol2019incremental}. For simplicity and comparison, we denote VINS-Mono as {P} which only uses point features, the method in \cite{rosinol2019incremental} as {PP} which uses point features and co-planarity constraints, and PL-VIO as {PL} which uses both point and line features. Our method is denoted as {PLP} which uses point, line and co-planarity constraints. For the effectiveness of the ablation study on heterogeneous features, we reimplement PP with VINS-Mono so that P, PP, PL, and PLP are all extended from the VINS-Mono code. The same parameters configuration are also used. We chose the absolute pose error (APE) as the main evaluation metric, which directly compared the trajectory error between the estimate and the ground truth \cite{sturm2012benchmark}. We also use relative pose error (RPE) to further obtain the qualitative analysis. The \texttt{TUM} evaluation toolkit \cite{sturm2012benchmark} was used to calculated APE and RPE in our experiments. 
      
\subsection{Synthetic Data}

\begin{figure}[thpb]
	\centering
	\includegraphics[width=0.4\textwidth]{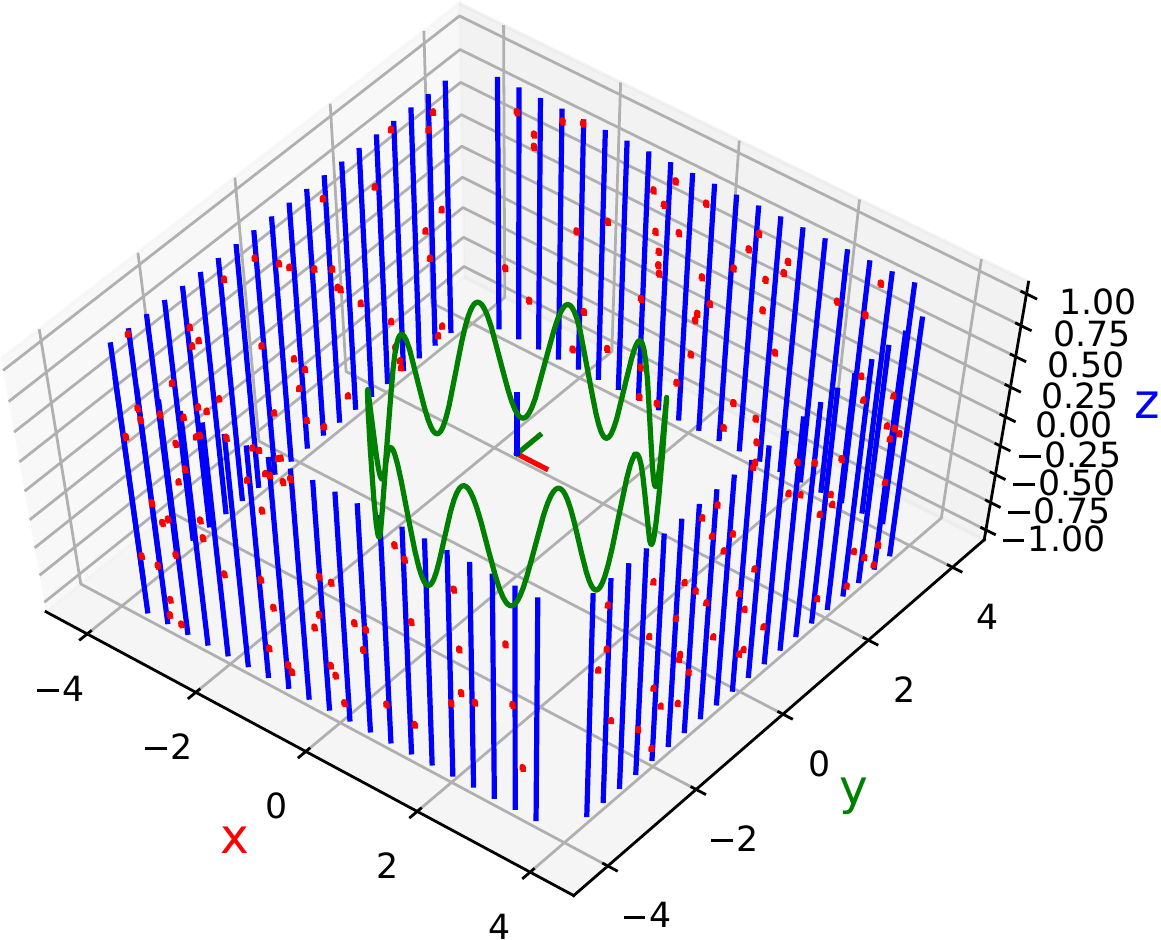}
	\caption{A Synthetic scene with line features (blue), point features (red) and camera trajectory (green).}
	\label{Synthetic}
\end{figure}

To verify the validity of VIO estimator with planar constraint, we generated a synthetic environment in which line and point features are distributed on the walls of a square room which side length is $8$ meters, as shown in Fig. \ref{Synthetic}. The virtual camera moved in the room, collecting images of 640 $\times$ 480 pixels. Nearly 15 point features and 8 straight lines are observed in each frame. Gaussian white noises with standard deviation $\delta_{px} = 1 $ pixel were added to the points and endpoints of lines in the captured images. 

Table \ref{simulres} summarizes the results of the methods {P}, {PP}, {PL} and the proposed {PLP} on the synthetic data. The root mean squared error (RMSE) of APE was used to evaluate the accuracy. Map error means the root mean square error of distance between true 3D position and the position we estimated. The results showed co-planarity constraints could improve the trajectory accuracy and our proposed {PLP} outperformed the comparison methods. For mapping quality, it can be seen that the largest map error from PL is caused by the inaccurate triangulation of the straight line endpoints. However, plane constraint can effectively improve the accuracy of the straight line and the mapping results.


\begin{table}[htpb]
	\caption{Results for pipelines {P}, {PP}, {PL} and our proposed aprroach {PLP} on the synthetic data. In \textbf{bold} the best result.}
	\label{simulres}
	\begin{center}
		\begin{tabular}{c c c c c}
			\hline
			& P & PP & PL &PLP\\
			\hline
			RMSE APE (cm)	& 8.94 & 8.06 & 7.97 & \textbf{7.34} \\
			Map error (cm) 	& 3.21 & 1.57 & 3.44 & \textbf{1.32} \\
			\hline
		\end{tabular}
	\end{center}
\end{table}

\subsection{EuRoc Dataset}

The EuRoc micro aerial vehicle (MAV) datasets were collected by a MAV in two indoor scenes, which contain stereo images from a global shutter camera at 20FPS and synchronized IMU measurements at 200 Hz. The two scenes contain walls, floor, and some other planar objects. Each dataset had a ground truth trajectory and a ground truth point cloud which can be used to evaluate localization accuracy and the 3D mesh map quality. In our experiments, the extrinsic and intrinsic parameters are specified by the datasets. Besides, we only used the images of the left camera. 

\subsubsection{Localization Accuracy}

we evaluated the RMSE APE of pipelines OKVIS, {P}, {PP}, {PL} and {PLP} on EuRoc dataset. Tab. \ref{table:eurco} shows the results of different methods. OKVIS results are from \cite{he2018pl}. For the experiment details of {P}, {PP}, {PL} and {PLP}, we use the default parameters from \cite{qin2018vins}, except the number of features. To improve the effectiveness of mesh detection, we increase the number of point features that each new frame image will collect 200 point features and all line features with its length more than 50 pixels. 

From the Tab. \ref{table:eurco}, PLP can improve the performance of PL on all sequences, but PP is not necessarily effective for P. This shows that using only sparse point features to detect the plane can sometimes bring incorrect structural information, and the introduction of straight lines can improve the accuracy of plane detection. 
In addition, Our approach using all point, line and plane features achieves the smallest translation error on all sequences except for \texttt{V1\_02}. 
The sequences of \texttt{V1} were collected in a square room with rich textures on walls and floors. 
Further analysis of the results on \texttt{V1\_01} and \texttt{V1\_02} shows the texture is so rich that P with only point features has achieved high accuracy and mismatched straight line features in PL will reduce the accuracy. However, since the camera movement speed in the \texttt{V1\_03} is very fast compared to \texttt{V1\_01} and \texttt{V1\_02}, the blurring and illumination-changing caused by high-speed motion will decrease the accuracy of P while the straight line features matching and detection are more stable so that the accuracy of PL is higher than P. The structure regularities of plane are valid for P and PL on all the sequences of \texttt{V1}.

\begin{table}[t]
	\caption{The RMSE of the state-of-art methods compare to our PLP method on \texttt{EuRoc} dataset. The translation (cm) and rotation (deg) error are list as follows. In \textbf{bold} the best result. }
	\label{table:eurco}
	\begin{center}
		\begin{tabular}{p{0.7cm} p{0.3cm}p{0.3cm} p{0.3cm}p{0.3cm}  p{0.3cm}p{0.3cm} p{0.3cm}p{0.3cm} p{0.3cm}p{0.3cm} p{0.3cm}p{0.3cm}}
			\hline
			\multirow{2}{*}{Seq.}
			& \multicolumn{2}{c}{OKVIS} & \multicolumn{2}{c}{P} & \multicolumn{2}{c}{PP} & \multicolumn{2}{c}{PL} & \multicolumn{2}{c}{PLP}\\ \cline{2-11}
			& \footnotesize{trans.}	& \footnotesize{rot.} 
			& \footnotesize{trans.}	& \footnotesize{rot.}
			& \footnotesize{trans.}	& \footnotesize{rot.}
			& \footnotesize{trans.}	& \footnotesize{rot.} 
			& \footnotesize{trans.}	& \footnotesize{rot.}\\
			\hline
			\texttt{MH\_02} & 36.0 &3.5 & 15.3 & 2.1 & 14.4 &2.0 & 11.7 &1.8 & \textbf{10.9} &\textbf{1.7} \\
			\texttt{MH\_03} & 21.5 & 1.5 & 18.5 & 1.3 & 18.6 & 1.3 & 15.4 & \textbf{1.1} & \textbf{14.9} & \textbf{1.1} \\
			\texttt{MH\_04} & 24.0 & \textbf{1.1} & 24.7 & 1.7 & 24.5 &1.7& 21.6 & 1.5 & \textbf{20.6} & {1.4}  \\
			\texttt{MH\_05} & 39.6 & \textbf{1.1} & 22.2 & 1.3 & 22.3 & 1.3 & 18.3 & 1.2 & \textbf{16.6} & 1.2   \\
			\texttt{V1\_01} & 8.6  & 5.5 & 6.1 & \textbf{5.2} & 5.7 & \textbf{5.2} & 6.4 	& 5.7 & \textbf{5.4} & \textbf{5.2}      \\
			\texttt{V1\_02} & 12.2 & 2.3 & 7.5 & \textbf{1.8} & \textbf{7.4} & \textbf{1.8} & 7.6 & \textbf{1.8} & 7.5 & \textbf{1.8} \\
			\texttt{V1\_03} & 19.6 & 3.8 & 12.8 &4.4 & 12.3 & 4.0 & 11.4 & 3.8 & \textbf{11.0} & \textbf{3.3}       \\
			\texttt{V2\_02} & 18.2 & 2.7 & 15.7 & 3.2 & 13.7 & 2.6 & 15.0 & \textbf{2.4} & \textbf{12.5} & \textbf{2.4}  \\
			\texttt{V2\_03} & 30.5 & 4.3 & 20.0 &3.4 & 16.8 &3.4& 16.6 &3.2& \textbf{15.4} & \textbf{2.4} \\
			\hline
		\end{tabular}
	\end{center}
\end{table}

\begin{figure}[t]
	\centering
	\includegraphics[width=\columnwidth]{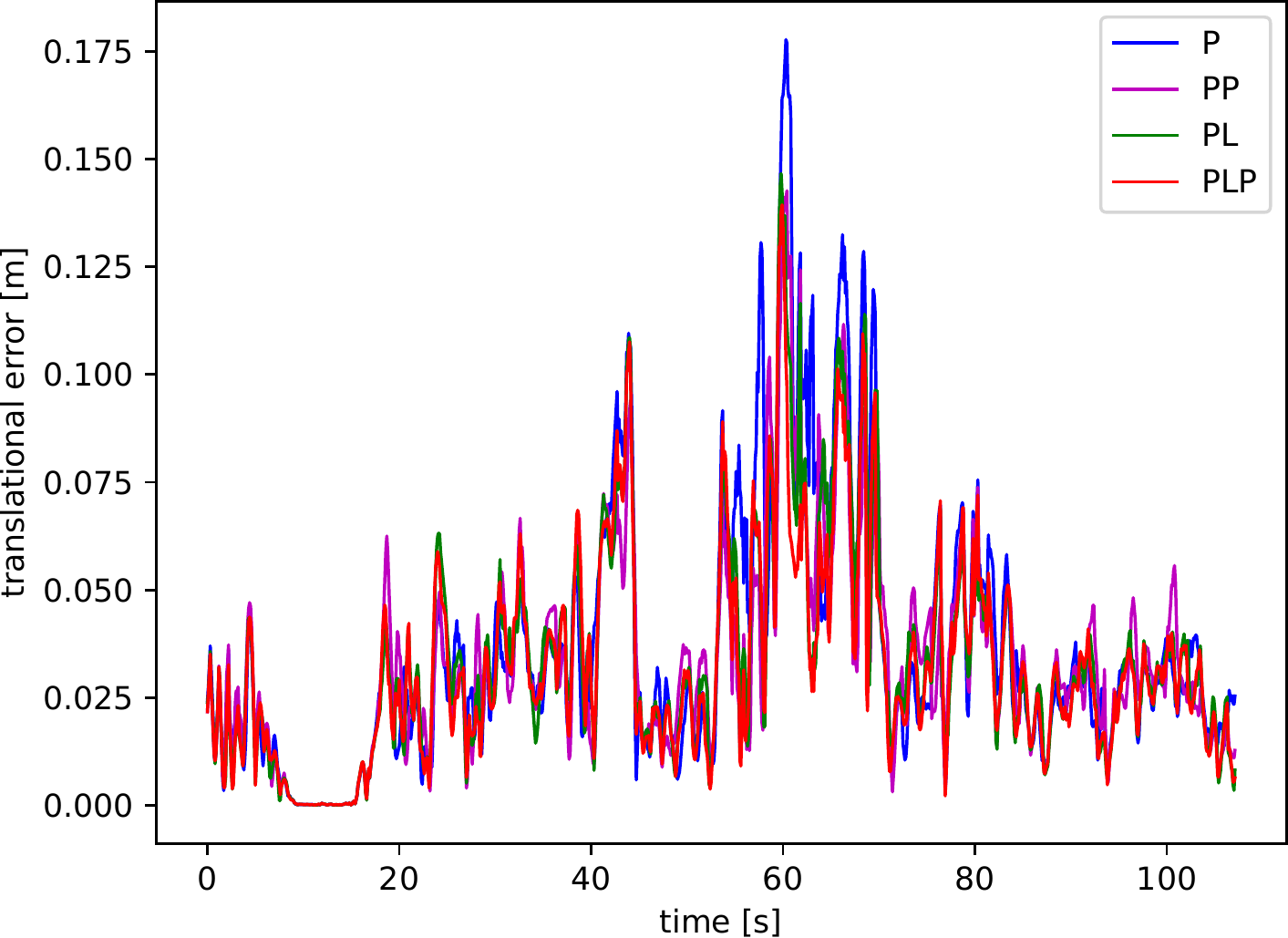}
	\caption{Translation error of RPE on dataset \texttt{MH\_05} with the approaches P, PP, PL, PLP (proposed).}
	\label{fig:rpe}
\end{figure}
\begin{figure}[thpb]
	\centering
	\includegraphics[width=0.8\columnwidth]{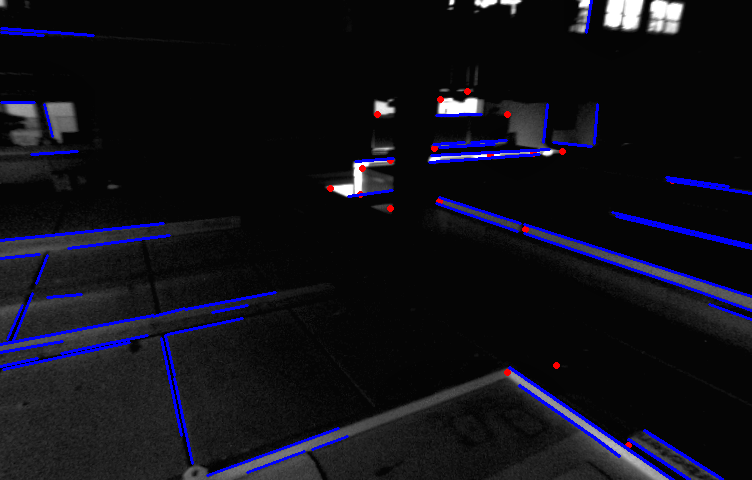}
	\caption{An example of the scene with rich line features and few point features on a plane. Detected points and lines are drawn in red and blue, respectively.}
	\label{fig:rpe_img}
\end{figure}

To get more insight into the effectiveness of our system, RPE is used to investigate the local accuracy of a trajectory. Fig.\ref{fig:rpe} shows the results on \texttt{MH\_05}. PLP presented the lower translation error of the RPE than the other methods, especially within the time range $57\sim67$s. An image example recorded within this time range is shown in Fig.\ref{fig:rpe_img}, which demonstrated a scene with few point features and many structural lines on a plane.
Compared with point features, the use of line features or plane structural regularities can effectively improve the positioning accuracy when point features cannot be effectively used for localization. In addition, the accuracy of PP is sometimes lower than P, such as around $t = 20$s and $t = 90\sim100$s. In comparison, straight lines make the plane detection more accurate, and the PLP error is smaller or closer than PL. Thus it verified the validity of the straight line in retaining spatial structure information.


\subsubsection{Mapping Quality}
\begin{figure}[thpb]
	\centering    
    \includegraphics[width=.9\columnwidth]{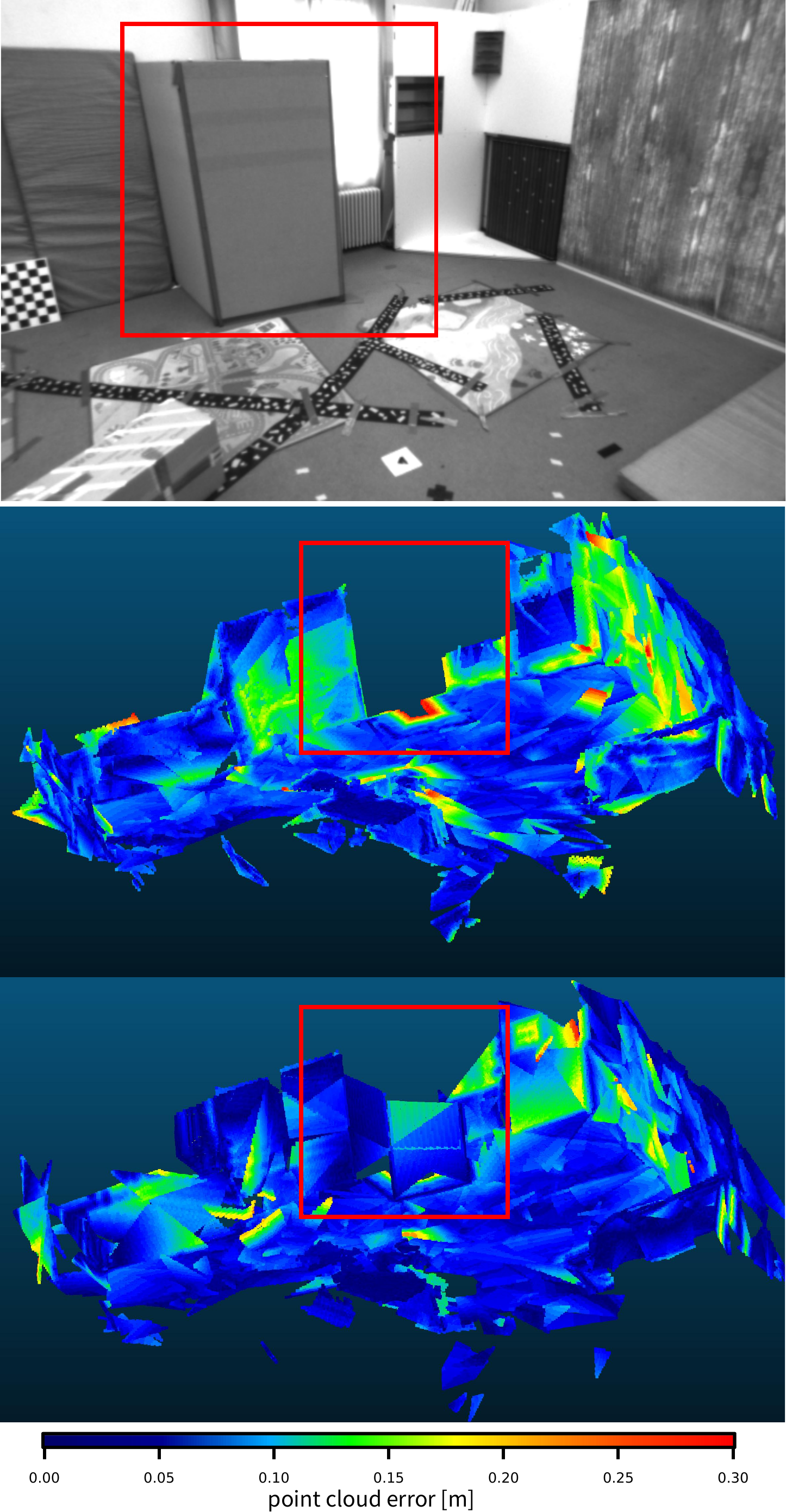}
	\caption{Mapping quality comparison. The two colorful point clouds are generated by {P} (middle) and {PLP} (bottom) running on \texttt{V1\_01}. Colors encode the corresponding mapping errors. The large box with textureless surfaces (red rectangle) (top) can be reconstructed by PLP.}  
	\label{fig:map_quality}     
\end{figure}

To evaluate the methods based on the metric of mapping quality, which is defined in \cite{schops2017multi}, we converted the generated a 3D mesh to a dense point cloud by sampling the mesh with a uniform density of $10^3$ points/$m^2$ \cite{rosinol2019master}. The resulting point cloud is aligned with the ground truth point cloud using the Iterative Closest Point (ICP) algorithm. For each point, the mapping error was its distance to its nearest ground truth point. To further verify that the geometric structure of the straight line is effective for plane detection and dense reconstruction, and to verify that the plane structure can improve the accuracy of the map, we compared the PLP with P. For the method P, we use the sparse point features to construct the mesh for dense reconstruction, but do not use structural constraints in optimization. The results given by P and PLP on dataset \texttt{V1\_01} are shown in Fig.\ref{fig:map_quality}. PLP gave a more complete dense map than P because the straight line gave better structural information during the mesh reconstruction process, especially for objects with the textureless surfaces, such as the large box object marked by the red line in Fig.\ref{fig:map_quality}. Moreover, the PLP has a lower mean distance error ($2.8$cm) than P ($3.6$cm) benefit from the planar regularities.

\subsubsection{Runtime Evaluation}

Because line and plane were added in the sliding window optimization, the state vector's size and the problem complexity are both increased. Here, we mainly compared the average execution time of the back-end module with other methods', which are evaluated on the \texttt{V1\_02} sequence. From Tab.\ref{tab:execution time}, we can observe that our non-iterative method detected plane and created mesh efficiently and the runtime was under $2$ms. The time cost of the optimization and marginalization was increased by only a few milliseconds. The reason was that Schur complement trick was used to accelerate the solver, besides, the plane number ($\approx 5$) and line number ($\approx 30$) were fewer compared with point number ($\approx 200$) in a sliding window.
\begin{table}[h]
	\caption{Mean execution time (Unit: millisecond) of different approaches run with the sequence \texttt{V1\_02}.}
	\label{tab:execution time}
	\centering
	\begin{center}
		\begin{tabular}{c c c c c}
			\hline
			Module  & P & PP & PL & PLP\\
			\hline
			\texttt{plane detection} & 0     & 0.39  & 0     & 0.44 \\
			\texttt{Mesh Creation}   & 0     & 0.65  & 0     & 1.42  \\	
			\texttt{Optimization}    & 31.61 & 33.63 & 30.59 & 35.88 \\
			\texttt{Marginalization} & 6.05 & 6.27 & 7.29 & 8.34  \\	
			\hline
		\end{tabular}
	\end{center}
\end{table}

\section{CONCLUSIONS}

In this paper, we present a novel VIO system combining point, line and plane features, which is called PLP-VIO. Both points and structural lines are used to detect plane and build 3D mesh, which is superior to the detection with only sparse points. The integration of planar regularity and heterogeneous visual features can significantly improve the localization accuracy and dense mapping accuracy simultaneously.


In the future, we intend to combine our system with deep learning to extract features and estimate dense depths. Also, we plan to embed the loop-closure module in our VIO system to build a globally consistent dense map for robot navigation applications.

\section*{ACKNOWLEDGMENT}
We would like to acknowledge insightful thoughts from Ji Zhao. We also thank Fangbo Qin for his suggestions on scientific writing.

\addtolength{\textheight}{-12cm}   

\bibliographystyle{ieeetr}
\bibliography{myref}

\end{document}